# Provably Optimal HBT Legalization in Seconds: Certified Optima Reveal Where the Headroom Really Lies

Kadhir Parani[1,$], Manjunath Bezawada[1], Divya M[1], Kesana Jathin[1], Kolli Sarath[1,*], Sm Zobaed[2,*]

[1] Boltzmann Labs, Hyderabad, India

[2] Department of Computer Science & Engineering Technology, University of Maryland Eastern Shore, Princess Anne, MD 21853, USA

[$] Present address: Department of Agricultural & Biological Engineering, Purdue University, West Lafayette, IN, USA

[*] *Corresponding authors*

kparani@purdue.edu; kollisarath@boltzmann.co; szobaed@umes.edu

**Abstract—**Hybrid bonding terminal (HBT) legalization assigns every terminal of a face-to-face 3D IC to its own pitch-grid site with minimum total Manhattan displacement. Open3DBench reports that a reinforcement-learning (RL) legalizer reduces displacement by 19–22% relative to a sliding-window (SW) heuristic, but none of its legalizers is publicly released. We ask how much displacement any legalizer can still recover. We build an open benchmark of 21 design/pitch instances from Open3DBench placement data, with reconstructed greedy, SW (ViaLegal-style) and BTAssign-style baselines, and GEO-Exact, an exact solver that runs successive shortest augmenting paths over lazily generated spatial candidate edges. An independent dual certificate proves its assignment optimal on all 21 instances (relative primal–dual gap at most $8.2\times10^{-16}$). Against these optima, our SW reconstruction with ViaLegal's default configuration is on average 0.011% above optimal (worst 0.119%), while on the 16 contended instances BTAssign-style, a density-guided heuristic (DGHL) and greedy are 3.0%, 6.9% and 64.5% above. Headroom does appear outside this regime: with 2×2 windows SW reaches 34.1%, and above about 58% utilization, on three of the six designs we swept, it reaches 21.1% on bp, 27.5% on swerv and 8.9% on jpeg in pitch sweeps (217 further certified instances, and ten bp points checked against an independent exact solver). If the published 19–22% gain were measured on instances like ours, SW would have to be at least 24–28% above optimal on average, far more than any SW variant we tested at nominal pitch; the RL paper's own comparison with the Hungarian optimum puts its SW 3.05% above optimal on average. GEO-Exact is also about an order of magnitude faster than DGHL and SW in our harness (geometric means 8.1× and 40×; a compiled core against Python baselines on a shared server). Code, optimal values and certificates are released.

## 1 Introduction

In face-to-face (F2F) 3D integration, signals cross between stacked dies through hybrid bonding terminals (HBTs), fine-pitch pads that must sit on a regular grid. A 3D placer produces ideal terminal positions that are not on that grid, so a legalization step gives every terminal its own grid site while keeping the total displacement small. In plain words it is a seating problem: many terminals want the same nearby seats, and the best arrangement may need a chain of small moves rather than one large one.

Open3DBench [1] is an open benchmark for 3D-IC backend implementation. Its Table V compares a greedy nearest-site legalizer, a sliding-window (SW) legalizer attributed to ViaLegal [2] and an RL legalizer [3], and reports that RL reduces displacement relative to SW by 22.06% at 4 µm pitch and 19.23% at 6 µm (averages over seven designs). The public repository (release v1.1, July 2026) contains placement data and HBT models, but we found none of the three HBT legalizers in it, so these results cannot be reproduced or extended.

We ask a question that does not depend on any particular heuristic: **how much displacement can any legalizer still recover?** The problem is a bipartite assignment, so its optimum can be computed and certified, and every method's gap to it measured rather than bounded. The RL paper [3] already compares its SW and RL with a Hungarian optimum on three public designs at 3–7 µm and one industrial design. We extend this to certified optima on full instances of up to 39,814 terminals, computed in seconds, and use them to test how much of the published gain is headroom. Our contributions are:

- **An open benchmark harness** of 21 design/pitch instances built from Open3DBench placement data, with reconstructed greedy, SW (ViaLegal-style) and BTAssign-style baselines and a fast density-guided heuristic (DGHL), all checked by one independent legality and cost checker (Section 4).
- **GEO-Exact**, an exact solver whose output on all 21 instances is proven optimal by an independent dual certificate (Sections 3 and 5).

- **A headroom analysis.** At the nominal pitches our SW reconstruction is within 0.119% of the optimum on all 21 instances, but the margin depends on SW's configuration and can vanish at high utilization, where SW is up to 27.5% above optimal. The optima also put the published RL gains in context (Section 7.8).

The paper does not compare against RL, for which no implementation is available, and its baselines are reconstructions from published descriptions. It evaluates one objective, total L1 displacement, and no downstream routing or timing metric.

## 2 Related Work

**Terminal assignment for 3D ICs.** ViaLegal [2] legalizes bonding terminals by minimum-cost bipartite matching inside a sliding window, because the full problem was considered too large; its default window is 3×3 tiles with stride 2 and a tile of 30× the via pitch. BTAssign [4] uses hierarchical bipartite matching over a quadtree partition. Net-to-pad assignment [5], unified 3D flows [6] and heterogeneous routing [7] treat terminals within broader flows, and the RL legalizer [3] starts from a greedy assignment and learns which subregions to refine; against a Hungarian optimum on ariane, bp and swerv_wrapper at 3, 5 and 7 µm and one industrial design, it reports SW 0.23–17.93% above optimal (3.05% on average) and RL 0.13%. Cell-level legalization [8, 9, 16, 17] and 3D placement [10] produce the ideal positions that terminal legalization consumes.

**Exact assignment.** Minimum-cost bipartite matching is solved exactly by the Hungarian method [11] and by shortest augmenting paths with dual potentials [12, 13]; network-flow [14], auction [18] and cost-scaling [19] methods are alternatives, and geometric matching avoids materializing all edges [15]. We do not claim a new assignment algorithm. The contribution is an exact implementation for this problem, an independent certificate of its output, and what the optima reveal about heuristics.

## 3 Problem and Optimality Certificate

Given $n$ ideal positions $p_1, \dots, p_n \in \mathbb{R}^2$ and $m \geq n$ legal sites $c_1, \dots, c_m$ given by the pitch and the placeable row area, legalization seeks an injective assignment $f$ minimizing

$$\min_f \; \Sigma_{i=1}^{n} \, \| p_i - c_{f(i)} \|_1 \tag{1}$$

With costs $c_{ij} = \|p_i - c_j\|_1$ this is a rectangular assignment problem (Figure 1). Its linear-programming relaxation has an integral optimum, and its dual is

$$\max \; \Sigma_i \, u_i - \Sigma_j \, w_j \;\; \text{s.t.} \;\; u_i - w_j \leq c_{ij} \; (\forall \, i, j), \;\; w \geq 0 \tag{2}$$

Any feasible $(u, w)$ gives a lower bound on the optimum, so a legal assignment whose cost equals a dual value is optimal. Given $u$, the best feasible $w$ is $w_j = \max(0, \max_i(u_i - c_{ij}))$. Our certificate takes only the terminal potentials $u$ from the solver, computes every $w_j$ by an exhaustive scan over all terminal–site pairs, and checks that the dual value equals the primal cost. It therefore does not rely on the solver's own claim of optimality.

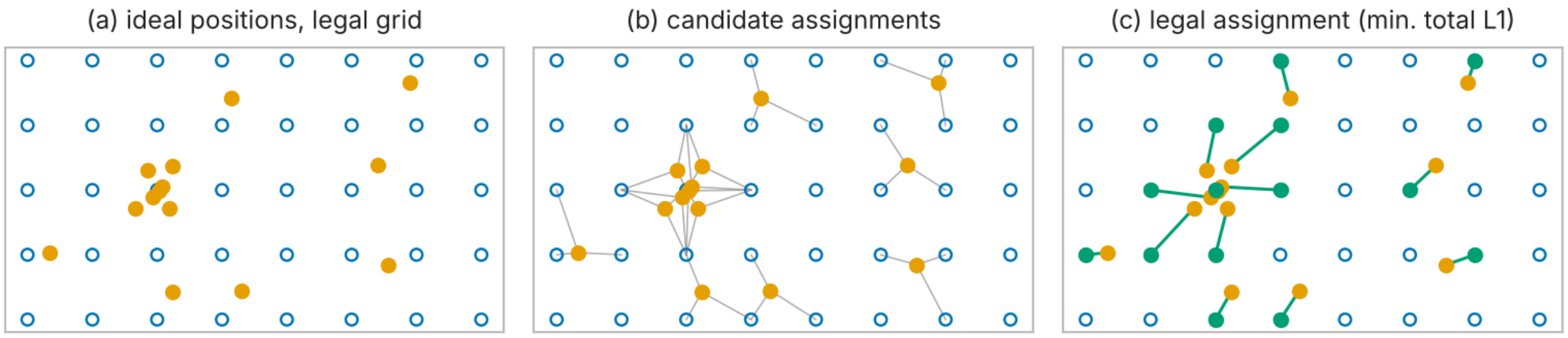

**Figure 1.** The problem on a toy instance (14 terminals, 8×5 grid). (a) Ideal positions (orange) and legal sites (blue). (b) Candidate assignments to the three nearest sites. (c) The assignment with minimum total L1 displacement (computed exactly).

## 4 Benchmark

### 4.1 Instances

Ideal HBT positions are extracted from placements shipped with Open3DBench, using the coordinate scaling of its codebase: for the Train, Test A and bp_quad designs (ICCAD 2023 benchmarks) the output of the TCAD'25 placer, and for the Test B designs (ICCAD 2022 benchmarks) the placement whose terminal counts match Open3DBench's cadb1021 rows. Legal sites come from each design's DEF floorplan, with the placement-row bounding box as placeable region, at a chosen pitch. Table 1 lists the 21 instances: four Train and three Test A designs at 4 and 6 µm, the large design bp_quad at 4 and 6 µm, and five Test B designs at their 2 µm pitch. The split exists because DGHL was tuned on Train only (Appendix A). We call the 16 Train, Test A and bp_quad instances *contended*: there, terminals compete for sites; at 2 µm the Test B terminals do not (Section 7.4).

**Table 1.** Designs: terminal count *n* and HBT utilization *n*/*m* (%) at the pitches used. Site counts are in Table B1.

| Design | Split | n | Util. 4 µm | Util. 6 µm | Util. 2 µm |
|---|---|---|---|---|---|
| ariane133 | Train | 4,934 | 8.1 | 18.1 | – |
| bp | Train | 7,667 | 15.5 | 34.8 | – |
| bp_be | Train | 2,702 | 15.0 | 33.3 | – |
| swerv_wrapper | Train | 5,755 | 16.9 | 37.9 | – |
| ariane136 | Test A | 4,793 | 7.8 | 17.6 | – |
| bp_fe | Test A | 2,790 | 20.3 | 45.7 | – |
| bp_multi | Test A | 7,151 | 18.2 | 41.0 | – |
| bp_quad | bp_quad | 39,814 | 10.3 | 23.1 | – |
| aes | Test B | 1,492 | – | – | 5.5 |
| dynamic_node | Test B | 3,083 | – | – | 54.8 |
| ibex | Test B | 1,600 | – | – | 28.4 |
| jpeg | Test B | 13,458 | – | – | 59.8 |
| swerv | Test B | 11,285 | – | – | 36.8 |

**Relation to Open3DBench.** The terminal counts equal those of Open3DBench's Table III (TCAD'25 rows, seven Train and Test A designs) and Table VI (cadb1021 rows, five Test B designs); for bp_quad, which those tables do not list, the count equals the NumTerminals field declared in the raw placement file. The legal regions differ, however: Table V's HBT utilization is 1.03–1.23× ours on all seven designs at both pitches (ariane133: 9.00% and 20.32% there, 8.05% and 18.12% here), so Table V used fewer sites. Runtimes also differ in scale (ariane133 SW: 2,982 s and 763 s in Table V, 3.4 s and 4.2 s per call here). Table V's SW displacement values (reported in $10^{-3}$ m) are also 7–16× smaller than our SW totals for the same design and pitch, although Table V uses fewer sites; the paper does not define the quantity precisely enough for us to reconcile this. We therefore compare methods only with each other on identical inputs, never with Table V's absolute values. The aes data exist in two floorplan directories that differ only in letter case; we use the larger (27,060 sites at 2 µm).

## 4.2 Baselines and checking

**Greedy** assigns terminals in turn to the nearest free site (expanding search over a spatial hash). **SW** tiles the design (tile 30× the pitch), solves an exact local assignment in each 3×3-tile window with a stride of two tiles, and lets earlier terminals be reassigned across overlapping windows; these are ViaLegal's stated defaults [2]. **BTAssign-style** matches over two levels of overlapping quadtree windows with improvement-only reassignment across boundaries. **DGHL** ranks coarse cells (30× the pitch) by the ratio of terminals to sites, solves bounded windows (at most 576 terminals) exactly in that order and places larger windows by a nearest-free-site fallback, then refines in two more passes with shrinking windows (Appendix A). Greedy, SW and BTAssign-style are our reconstructions, since the original implementations are not public; DGHL is our own. All four are Python with NumPy and SciPy.

One shared checker validates every assignment in this paper: all site indices are in range, no site is used twice, the count equals *n*, and the total recomputed from scratch equals the reported one. It rejects four negative controls (duplicate site, index out of range, wrong count, cost off by $10^{-6}$).

# 5 GEO-Exact

**The idea.** Each terminal wants its nearest free site; when two want the same one, the cheapest fix is a chain of moves. Classical exact solvers find the cheapest chain in a graph that joins every terminal to every site, which is far too large. GEO-Exact searches the same graph but creates an edge only when the search could still need it: sites are stored in a uniform grid, every ring of cells around a terminal has a known minimum distance, and a ring is opened only when that bound could still improve the best path found so far.

**Algorithm.** GEO-Exact keeps a partial matching and dual potentials $u_i$ (terminals) and $v_j$ (sites) with nonnegative reduced costs $c_{ij} - u_i - v_j$, tight matched edges and zero potential on free sites. Terminals are inserted one at a time; a heap-based Dijkstra search over reduced costs finds the shortest augmenting path to a free site, and potentials are updated to keep the invariants. The search's edges are never stored. Sites are bucketed into a grid with cell size equal to the pitch, and opening a ring (or, in the default variant, a block of cells) is itself a heap event keyed by a lower bound on the reduced cost of every site it contains. When the target of the shortest path is finalized, every unopened edge has a reduced cost at least as large, so none can give a cheaper path; this is the exactness argument the code implements. Terminals are inserted sparse-first (by a local density count); the order changes the work and which optimum is returned among ties, not the optimal value. The core is C (GCC -O3) called from Python, and no $n \times m$ matrix is stored.

**Evidence of exactness.** (i) The certificate of Section 3 passes on all 21 instances; the adapter that exports $u$ returns assignments identical to the solver's. (ii) The ring and block variants agree on all 21 instances (difference at most $1.5\times10^{-11}$). (iii) GEO-Exact equals dense Hungarian over all sites on 28 real 300-terminal subsets. We make no worst-case complexity claim; the running time depends on how many events the geometry generates (Section 7.6).

## 6 Experimental Setup

Experiments ran on a shared Linux server (Intel Xeon W-2133, 6 cores / 12 threads, 61 GB, no GPU) with Python 3.13.11, NumPy 2.5.2 and SciPy 1.18.0. The *gap* of a method is its total displacement divided by the certified optimum, minus one, in percent; aggregates are means of per-case gaps. Runtime is the legalizer call only, in a fresh process, averaged over 5 trials per cell in the first timing run and 3 in the second; speed-ups are geometric means of per-case ratios, and a ratio within 1.2× of one is a tie. Displacement is deterministic: five trials on each of the 64 Train, Test A and bp_quad cells of the four heuristics gave identical totals and assignment hashes, and the released code reproduces all 56 Train and Test A totals of the four heuristics exactly. The server was never idle: load averages were 6–17 during timing, and single-case ratios moved by up to about 2× between runs (DGHL against GEO-Exact on ariane133 at 4 µm: 79× and 36× in two single-trial repeats, against a 5-trial mean of 46× in run 1). We therefore report two independent timing runs and read speed-ups as orders of magnitude.

## 7 Results

### 7.1 The optima are certified

The dual certificate passes on all 21 instances; the largest relative primal–dual gap is $8.25\times10^{-16}$ (bp_fe@6), i.e. floating-point rounding. The largest instance, bp_quad at 4 µm (39,814 terminals, 387,506 sites), has optimum 213,777.81 µm. Check times range from 0.6 s to 576 s (Table B1). The Test B pitch sweeps of Section 7.4 contain 222 certified instances, 217 of them beyond the five nominal ones. These certificates concern total L1 displacement on these inputs only.

### 7.2 Headroom at the nominal pitches

SW with ViaLegal's defaults equals the optimum (within $10^{-6}$%) on 17 of 21 instances and is above it on four, all at 6 µm: bp (0.119%), bp_quad (0.076%), bp_multi (0.024%), swerv_wrapper (0.009%). Its mean gap is 0.011% over all 21 instances and 0.014% over the 16 contended ones (Table 2). BTAssign-style is better than DGHL on all 16 contended instances and worse than SW on all 16; greedy is 34–102% above optimal there. At the nominal pitches, therefore, our SW reconstruction leaves essentially nothing to recover; other SW implementations can leave more (Section 7.8).

**Table 2.** Gap above the certified optimum (%), mean of per-instance gaps (number of instances in parentheses). Per-instance values: Table B1.

| Instances | SW | BTAssign-style | DGHL | Greedy |
|---|---|---|---|---|
| Train (8) | 0.016 | 2.998 | 7.288 | 64.193 |
| Test A (6) | 0.004 | 2.285 | 7.101 | 66.395 |
| bp_quad (2) | 0.038 | 4.773 | 5.130 | 60.129 |
| Test B, 2 µm (5) | 0.000 | 0.111 | 0.000 | 0.000 |

| Instances | SW | BTAssign-style | DGHL | Greedy |
|---|---|---|---|---|
| **Contended (16)** | **0.014** | **2.952** | **6.948** | **64.511** |
| All 21 | 0.011 | 2.276 | 5.294 | 49.151 |
| Worst instance | 0.12 (bp@6) | 9.20 (bp_multi@6) | 15.14 (bp_multi@6) | 101.99 (bp_multi@6) |

## 7.3 SW depends on its configuration

Varying one SW parameter at a time over all 21 instances (Table 3), larger windows, a smaller stride or larger tiles bring SW to the optimum (worst case at most 0.003%). Smaller tiles, no cross-window reassignment or 2×2 windows do not: the worst cases are 4.1%, 20.9% and 34.1%. ViaLegal searches window sizes 3–10 and strides 1–3 [2], so the 2×2 window lies outside its own range. Open3DBench describes its SW local solver as min-cost flow; replacing our local assignment solver by an independent exact matching solver (SciPy's min_weight_full_bipartite_matching) gave identical totals on 11 of 14 Train and Test A instances and differences of at most 0.004% (tie-breaking) on the other three.

**Table 3.** SW variants over all 21 instances: mean and worst gap above the optimum (%) and share of instances within 1% (%). One parameter differs from the default per row.

| SW variant | Mean gap | Worst gap | Instances ≤ 1% |
|---|---|---|---|
| **ViaLegal default (tile 30, window 3×3, stride 2)** | **0.011** | **0.119** | **100.0** |
| stride 1 | 1.8e-04 | 3.0e-03 | 100.0 |
| tile 40 | 1.3e-04 | 2.0e-03 | 100.0 |
| window 4×4 | 7.0e-06 | 1.5e-04 | 100.0 |
| tile 20 | 0.576 | 4.142 | 81.0 |
| no cross-window reassignment | 1.992 | 20.867 | 71.4 |
| window 2×2 | 6.643 | 34.057 | 28.6 |

## 7.4 Headroom appears at high utilization

Utilization $n/m$ rises with the pitch, because fewer sites cover the same area. We swept the pitch of bp from 6.9 to 9.9 µm (ten points) and of the five Test B designs from 2.0 µm in steps of 0.05 µm up to the last feasible pitch (222 feasible cells, including the five nominal Test B instances), certifying every Test B optimum; the ten bp optima were not dual-certified but agree with an independent sparse exact reference to within $5\times10^{-14}$% (Figure 2, Table 4, Table B3). Across all 248 distinct instances, our SW is at most 0.39% above optimal at or below 52% utilization (125 instances), and every gap above 1% occurs at 58.5% utilization or more.

On bp the gap rises to 21.13% at 64.5% utilization (8.2 µm), exceeds 5% between 58.5% and 69.5%, and falls to zero above 92%. On swerv (11,285 terminals) it reaches 27.5% at 90.8% utilization and exceeds 1% at 13 of 26 pitches; on jpeg (13,458 terminals) it reaches 8.9% at 93.5%. On aes, dynamic_node and ibex SW is optimal at every pitch, up to 95–100% utilization. In the Test B sweep DGHL is up to 37% and greedy up to 261% above optimal, and BTAssign-style raised an error in 8 cells (jpeg at 2.50–2.55 µm, swerv at 3.00–3.25 µm). We have not identified the mechanism, and no geometric measure we tried predicts which designs are affected. These utilizations are above those of Open3DBench's Table V (at most 49%) and of our nominal instances (at most 60%). They arise when the bonding grid is coarse relative to the number of crossing terminals, through a coarser pitch or more 3D connections per unit area; we use them as a stress test of windowed matching, not as a claim about current designs.

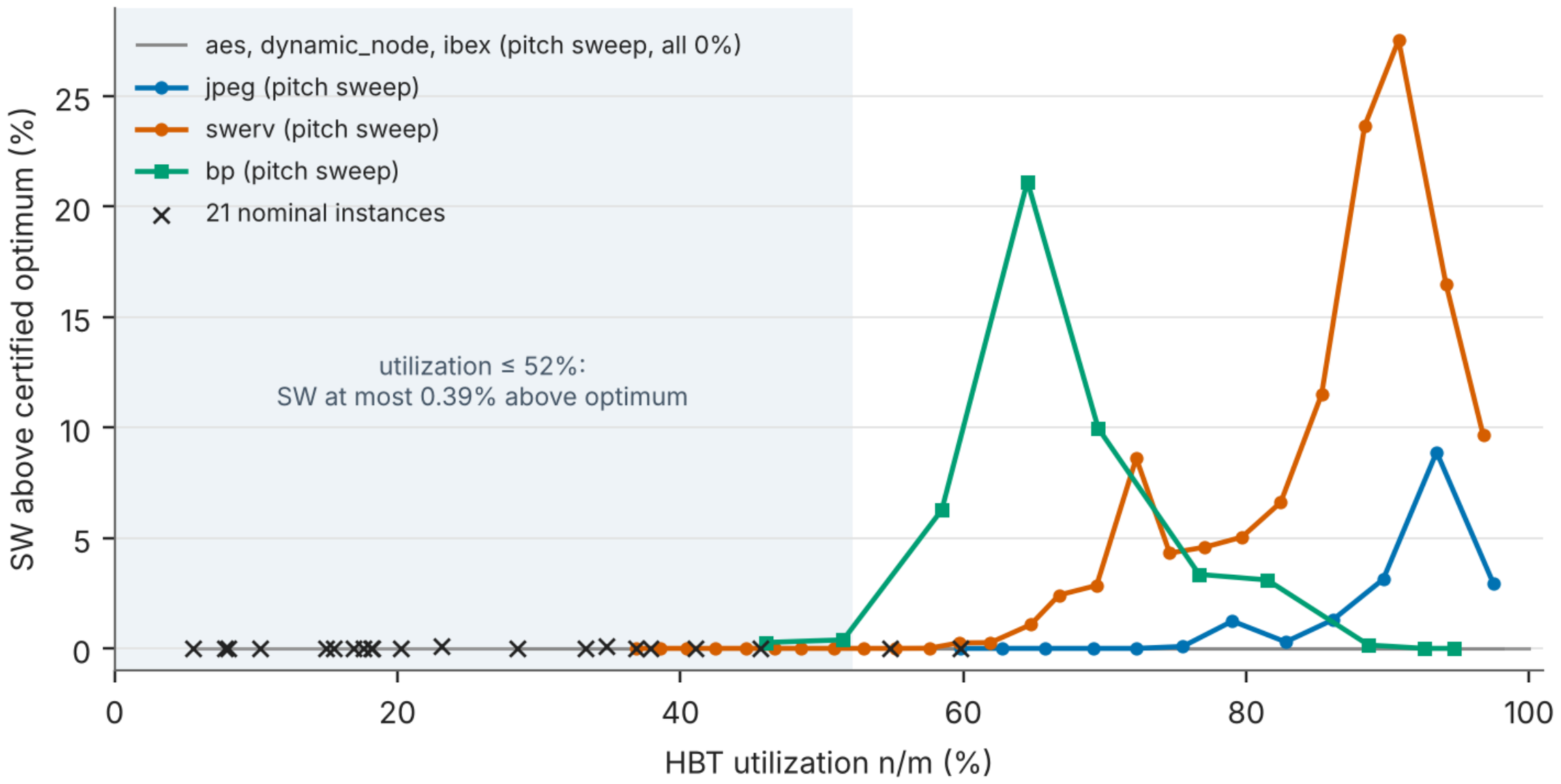

**Figure 2.** SW gap above the certified optimum against utilization, for the 21 nominal instances (crosses) and the pitch sweeps of bp and the five Test B designs. The 21 nominal and 217 further Test B optima are dual-certified; the 10 bp optima are GEO-Exact solutions checked against an independent sparse exact reference.

**Table 4.** Test B pitch sweep (step 0.05 µm): number of feasible pitches, pitch and utilization ranges, SW's worst gap (%, with its utilization), pitches with SW gap above 1% / 5%, and DGHL's worst gap (%).

| Design | n | Pitches | Pitch (µm) | Util. (%) | SW worst | SW > 1% / > 5% | DGHL worst |
|---|---|---|---|---|---|---|---|
| aes | 1,492 | 133 | 2.00–8.60 | 6–98 | 0.00 | 0 / 0 | 13.0 |
| dynamic_node | 3,083 | 14 | 2.00–2.65 | 55–95 | 0.00 | 0 / 0 | 26.1 |
| ibex | 1,600 | 37 | 2.00–3.80 | 28–100 | 0.00 | 0 / 0 | 20.0 |
| jpeg | 13,458 | 12 | 2.00–2.55 | 60–97 | 8.88 (93%) | 5 / 1 | 36.6 |
| swerv | 11,285 | 26 | 2.00–3.25 | 37–97 | 27.53 (91%) | 13 / 8 | 35.9 |

**Why Test B ties at 2 µm.** At the nominal pitch every method reaches the optimum on four Test B designs and BTAssign-style differs only on jpeg (+0.55%); the mean displacement is 0.85–1.01 times pitch/2, the value expected when every terminal gets its nearest site. The minimum Chebyshev spacing of the ideal terminals is exactly 2 µm in all five designs, but at most 1.3% of coordinates per axis lie on multiples of 2 µm (circular phase concentration at most 0.14, except 0.49 for dynamic_node's x coordinates). The source placement therefore follows a 2 µm minimum-spacing rule rather than a lattice, and contention-freedom at 2 µm is a property of the placement. The larger pitches are our extension, not the designs' own.

## 7.5 Speed

GEO-Exact is faster than DGHL on 17 of 21 instances, tied on 3 and slower on 1 (bp_fe at 6 µm, 0.66×), and faster than SW on all 21 (4.8–143×). Geometric-mean speed-ups are 8.1× over DGHL and 40× over SW in the first timing run (95% bootstrap intervals over instances [4.2, 14.7] and [26, 60]) and 7.5× and 37× in an independent second run (load 14–17; [3.8, 14.3] and [24, 55]), with the same faster/tied/slower counts (Table 5, Figure 3). The cases where GEO-Exact does not win are dense 6 µm instances. On bp_quad, GEO-Exact takes 0.65 s and 2.58 s at 4 and 6 µm, against 12.0 s and 6.4 s for DGHL and 37.4 s and 67.4 s for SW.

**Table 5.** Speed-up of GEO-Exact (geometric mean of per-instance runtime ratios; > 1 means GEO-Exact is faster) in two independent timing runs on a loaded server. Per-instance times of run 1: Table B2.

| Instances | Run 1 vs DGHL | Run 1 vs SW | Run 2 vs DGHL | Run 2 vs SW |
|---|---|---|---|---|
| Train (8) | 6.7 | 35.3 | 6.8 | 39.2 |
| Test A (6) | 4.7 | 27.7 | 3.8 | 22.4 |
| bp_quad (2) | 6.7 | 38.7 | 5.1 | 31.1 |

| Instances | Run 1 vs DGHL | Run 1 vs SW | Run 2 vs DGHL | Run 2 vs SW |
|---|---|---|---|---|
| Test B (5) | 22.5 | 80.9 | 23.4 | 67.2 |
| **All 21** | **8.1** | **40.5** | **7.5** | **37.1** |
| Faster / tied / slower | 17 / 3 / 1 | 21 / 0 / 0 | 17 / 3 / 1 | 21 / 0 / 0 |

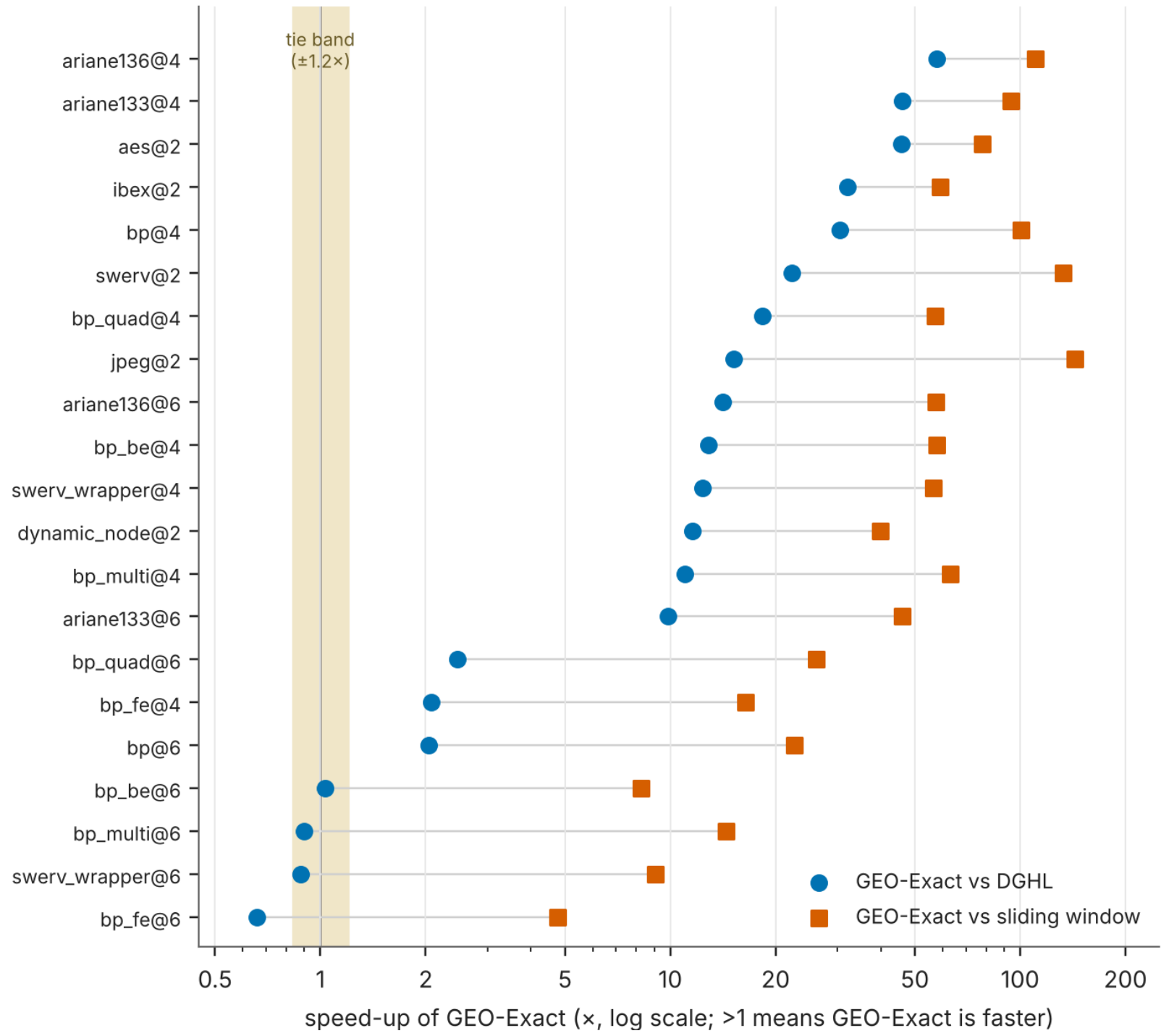

**Figure 3.** Per-instance speed-up of GEO-Exact over DGHL (circles) and SW (squares), run 1, log scale. The shaded band marks ties (within 1.2×).

**What the speed-up does and does not show.** GEO-Exact is compiled C and the baselines are Python, so part of the speed-up is language. One check bounds this. A compiled sparse exact solver (SciPy's LAPJVsp on a k-nearest-site graph, k from 8, doubled until a full matching exists) was less accurate than SW (mean +0.66%), needed k = 512 on bp_fe at 6 µm, did not finish within 900 s on swerv_wrapper at 6 µm and took over 400 s on bp_multi at 6 µm; GEO-Exact was about 9.6× faster while exact. A measurement on an idle machine remains to be done.

## 7.6 Scaling and adversarial inputs

On synthetic uniform layouts (up to 200,000 terminals) GEO-Exact's runtime grows as about $n^{1.05}$ at both utilizations (0.2 and 0.5); on clustered layouts (up to 30,000) the slopes are 1.71–1.82, similar to SW (1.68–1.83). These fits use four to six sizes. DGHL failed to place all terminals on every clustered run at 100,000 terminals and utilization 0.2 (four seeds, 3,076–4,977 terminals unplaced). GEO-Exact is not fast on every input: on hotspot instances (all terminals within 5% of the die width) it needed 11.7 million heap pops for 2,000 terminals and 109 million for 5,000 (about $n^{2.4}$, from two sizes), and a single row of 5,000 terminals needed 36.6 million. All of these adversarial results were certified optimal (hotspot with 5,000 terminals: 97 s to certify). Larger adversarial sizes were not completed, so worst-case behavior remains open.

## 7.7 Other objectives: L2 displacement and tails

The optima are for total L1 displacement, which is the objective Open3DBench states (cumulative Manhattan displacement [1]). Under Euclidean (L2) displacement the picture changes (Table 6). We compared each assignment's L2 total on the 16 contended instances with a sparse L2 solution (k-nearest-site graph, k = 32 doubled until feasible, solved exactly), which is an upper bound on the L2 optimum, so the excesses are lower bounds on the true L2 headroom. The L1 optimum is 2.25–11.03% above it (mean 5.07%), about the same as SW; on bp_fe at 6 µm DGHL is below both. The reason is ties: L1 optima are highly non-unique. Breaking L1 ties by minimum squared L2 reduces the L2 excess of the L1 optimum to 2.0% on average (maximum 4.6%) on 140 random 300-terminal patches. The conclusion that our SW leaves almost nothing to recover is therefore specific to the L1 objective.

**Table 6.** L2 total displacement above a sparse L2 solution (an upper bound on the L2 optimum), 16 contended instances (%).

| Assignment | Mean | Minimum | Maximum |
|---|---|---|---|
| GEO-Exact (L1 optimum) | 5.07 | 2.25 | 11.03 |
| SW | 5.15 | 2.38 | 11.35 |
| BTAssign-style | 7.79 | 2.55 | 14.23 |
| DGHL | 9.86 | 3.10 | 18.42 |
| Greedy | 70.99 | 40.61 | 106.89 |

Minimizing the total also does not minimize the tail. GEO-Exact's maximum displacement is no larger than DGHL's on all 16 contended instances (strictly lower on 15) and lower than SW's on 10, but higher than SW's on 5; its 95th percentile is higher than SW's on 12 of 16, lower on 3 and equal on 1 (Table B2).

### 7.8 What the optima imply for the published RL comparison

We cannot evaluate the RL legalizer, but the optima show what a gain over SW implies. If a method reduces displacement relative to SW by a fraction $r$, SW must be at least $r/(1-r)$ above the optimum. Open3DBench's averages of 22.06% (4 µm) and 19.23% (6 µm) would therefore require its SW to be at least 28.3% and 23.8% above optimal on average. On our instances, our SW reconstruction with ViaLegal's defaults is 0.011% above optimal on average over all 21 instances, and the weakest variant we tested (2×2 windows) 6.6%. The RL paper [3] measures its own SW against a Hungarian optimum (ariane, bp and swerv_wrapper at 3–7 µm, and one industrial design) and finds it 0.23–17.93% above optimal, 3.05% on average, with RL at 0.13%; this implies an average RL gain over SW of about 2.6%, far below Open3DBench's Table V averages. Its SW also takes 610–2,821 s per design, against seconds for ours, so it is configured differently from our reconstruction. Table V's utilizations (8.5–22% at 4 µm, 19–49% at 6 µm) lie below the 58% at which our SW first shows gaps above 1%, and its displacement values are not comparable with ours (Section 4.1). We therefore draw a conditional conclusion: with an SW configured as ViaLegal describes, on instances like ours, little displacement is left for any method to gain at nominal pitch. This says nothing against the published RL method, which may help against other SW implementations, at other utilizations or on downstream metrics; a controlled comparison needs Open3DBench's SW configuration, instances and RL code.

## 8 Limitations

- **Reconstructed baselines.** Greedy, SW and BTAssign-style follow published descriptions, not original code. The SW sweep brackets the unknown settings, but the RL paper's SW is 3.05% above optimal on its instances against 0.011% for ours on ours, so conclusions about headroom above SW apply to our reconstruction.
- **One objective.** The optimality claims are for total L1 displacement. We report no routing, timing or power effects; even an HPWL proxy was not possible, because neither the raw placement nor the DEF nets map the crossing terminals to nets.
- **Instances.** 21 instances from 13 designs, several sharing a lineage (ariane133/136, the BlackParrot variants); the high-utilization sweeps use pitches we chose. The large SW gaps are measured, but their mechanism is not identified.
- **Timing.** Measured on a loaded shared server, with a compiled solver against Python baselines; ratios are reliable to about a factor of two.
- **Scaling.** No worst-case complexity claim; hotspot inputs grow faster than linearly, and larger adversarial and clustered runs were not completed.
- **No RL comparison**, and a single technology (Nangate45) and machine.

## 9 Conclusion

Certified optima turn the question 'is there a better legalizer?' into a measurement. On 21 instances from Open3DBench data, our reconstruction of the sliding-window legalizer, configured as ViaLegal describes, is within 0.119% of the optimum everywhere, so at nominal pitches, against such a baseline, a new method can gain little displacement; the RL paper's own comparison (SW 3.05% above the optimum on average) shows that other SW implementations leave more. The headroom that does exist lies elsewhere: in weaker SW configurations (up to 34%), at utilizations above about 58% on some larger designs (up to 27.5%), and under objectives other than L1. GEO-Exact returns the certified optimum about an order of magnitude faster than the Python heuristics on most of these instances, though not on adversarial hotspots. Future work is a controlled comparison with Open3DBench's own SW and RL, downstream PPA evaluation, a predictor for when windowed matching fails, and timing on an idle machine.

**Availability.** The harness, baselines, DGHL, GEO-Exact, the instances' optimal values and certificates are available at [repository URL] (release [tag]; archive [DOI]). Whether the Open3DBench-derived instance files may be redistributed has not yet been confirmed; if not, the release contains the extraction scripts, which regenerate the instances from Open3DBench data.

## Appendix A DGHL

DGHL (Density-Guided Heuristic Legalization) has no learned component. It is included as a fast approximate baseline and as a test of whether scheduling exact local matching by contention helps.

```
Algorithm A1: DGHL
Input: ideal positions P (n), sites C (m >= n); passes = 3, k_hbt = 576, k_site = 10,000, sigma =
0.5
build k-d tree over C; bucket P and C into coarse cells (30 x pitch)
for t = 1..passes:
  rank cells by (terminals / max(sites, 1)), highest first
  window <- 90 x pitch * sigma^(t-1)
  for each cell in rank order:
    rows <- terminals currently in the cell's window (t = 1: unassigned only)
```

```
    candidates <- free-or-in-window sites in the window (at most k_site)
    if |rows| <= k_hbt: exact Hungarian on (rows, candidates);
                        for t > 1 keep it only if the window cost strictly decreases
    else if t = 1: nearest-free-site fallback
place any remaining terminals by the fallback; assert injective
```

**Behavior.** Instrumentation (identical assignments on the 16 contended instances) shows that 4–30% of windows exceed the 576-terminal cap but hold 60–100% of the terminals, which pass 1 places by the fallback; DGHL is therefore mostly a nearest-free-site placement followed by exact, improvement-gated refinement. It fails when its bounded fallback search finds no free site near a terminal: bp at 10 µm (96.8% utilization, 2,160 terminals unplaced) and the clustered synthetic runs of Section 7.6.

**Tuning.** The split was fixed before any search. Passes, k_hbt and the shrink factor were searched on Train only by a coordinate-wise automated search (k_site was fixed), gated by legality, runtime, memory and agreement with exact Hungarian on a 300-terminal subset. Three steps were accepted, all changing passes or k_hbt (passes 2→3, k_hbt 256→512→576); an alternative shrink factor was tried and rejected, so the accepted search changed two of the four hyperparameters. The steps lowered the Train objective (mean disadvantage relative to SW) from 0.2178 to 0.0727. DGHL's mean disadvantage relative to SW is 7.27% on Train and 7.10% on Test A. Tested at the design level (4 against 3 designs), the difference of 0.17 points has a 95% interval of [-10.4, 10.7], so this comparison cannot distinguish generalization from overfitting. Test A is held out by instance, not by design family.

**Ordering ablation.** Changing only the visiting order (contention, raster, 100 random permutations) on the 14 Train and Test A instances gives a mean disadvantage relative to SW of 7.195% for contention, 8.212% for random orders (SD 0.574 points) and 8.543% for raster. Contention beats 93 of the 100 random orders (permutation $p \approx 0.079$). Per instance, a sign test is significant on 11 of 14: contention wins on seven and loses on four, and neither pitch nor utilization predicts the direction. The ordering is worth about one point; the bounded-window refinement, common to all orders, accounts for most of DGHL's gap. Since GEO-Exact is exact and usually faster, DGHL's value here is as a measured example of what bounded-window heuristics cost.

## Appendix B Per-Instance Results

**Table B1.** The 21 instances: terminals *n*, sites *m*, utilization (%), certified optimum (µm), relative primal–dual gap, certificate check time (s), and the gap of each heuristic above the optimum (%).

| Instance | n | m | Util. | Optimum | Rel. gap | Check s | SW | BT-style | DGHL | Greedy |
|---|---|---|---|---|---|---|---|---|---|---|
| ariane133@4 | 4,934 | 61,256 | 8.1 | 19,471.02 | 1.9e-16 | 22 | 0.000 | 0.36 | 1.01 | 43.6 |
| ariane133@6 | 4,934 | 27,225 | 18.1 | 51,511.04 | 2.8e-16 | 10 | 0.000 | 0.16 | 5.01 | 52.8 |
| bp@4 | 7,667 | 49,506 | 15.5 | 24,858.73 | 0 | 27 | 0.000 | 0.40 | 5.31 | 43.9 |
| bp@6 | 7,667 | 22,052 | 34.8 | 103,511.77 | 2.8e-16 | 13 | 0.119 | 5.86 | 14.85 | 94.9 |
| bp_be@4 | 2,702 | 18,056 | 15.0 | 11,981.19 | 3.0e-16 | 3 | 0.000 | 0.52 | 4.10 | 66.0 |
| bp_be@6 | 2,702 | 8,118 | 33.3 | 67,910.17 | 2.1e-16 | 2 | 0.000 | 2.47 | 5.83 | 70.7 |
| swerv_wrapper@4 | 5,755 | 34,056 | 16.9 | 24,572.34 | 0 | 13 | 0.000 | 5.09 | 9.51 | 57.5 |
| swerv_wrapper@6 | 5,755 | 15,180 | 37.9 | 149,801.24 | 7.8e-16 | 10 | 0.009 | 9.13 | 12.68 | 84.2 |
| ariane136@4 | 4,793 | 61,256 | 7.8 | 15,319.95 | 0 | 20 | 0.000 | 0.03 | 0.69 | 33.6 |
| ariane136@6 | 4,793 | 27,225 | 17.6 | 44,668.69 | 1.6e-16 | 9 | 0.000 | 0.04 | 5.07 | 64.2 |
| bp_fe@4 | 2,790 | 13,775 | 20.3 | 31,949.09 | 3.4e-16 | 3 | 0.000 | 0.80 | 6.34 | 79.6 |
| bp_fe@6 | 2,790 | 6,111 | 45.7 | 123,512.27 | 8.2e-16 | 3 | 0.000 | 2.48 | 4.41 | 59.5 |
| bp_multi@4 | 7,151 | 39,204 | 18.2 | 31,816.96 | 1.1e-16 | 19 | 0.000 | 1.16 | 10.95 | 59.5 |
| bp_multi@6 | 7,151 | 17,424 | 41.0 | 151,740.50 | 3.8e-16 | 15 | 0.024 | 9.20 | 15.14 | 102.0 |
| bp_quad@4 | 39,814 | 387,506 | 10.3 | 213,777.81 | 8.2e-16 | 576 | 0.000 | 3.13 | 3.29 | 58.0 |
| bp_quad@6 | 39,814 | 172,225 | 23.1 | 676,286.39 | 1.7e-16 | 526 | 0.076 | 6.42 | 6.97 | 62.3 |
| aes@2 | 1,492 | 27,060 | 5.5 | 1,439.86 | 0 | 3 | 0.000 | 0.00 | 0.00 | 0.0 |
| dynamic_node@2 | 3,083 | 5,625 | 54.8 | 2,629.24 | 0 | 1 | 0.000 | 0.00 | 0.00 | 0.0 |
| ibex@2 | 1,600 | 5,625 | 28.4 | 1,537.78 | 0 | 1 | 0.000 | 0.00 | 0.00 | 0.0 |
| jpeg@2 | 13,458 | 22,500 | 59.8 | 13,537.65 | 0 | 22 | 0.000 | 0.55 | 0.00 | 0.0 |
| swerv@2 | 11,285 | 30,625 | 36.8 | 11,294.12 | 0 | 25 | 0.000 | 0.00 | 0.00 | 0.0 |

**Table B2.** Runtime of the legalizer call (s, run 1) with speed-ups of GEO-Exact, and maximum and 95th-percentile displacement (µm) on the contended instances.

| Instance | GEO s | DGHL s | SW s | × DGHL | × SW | GEO max | SW max | DGHL max | GEO p95 | SW p95 | DGHL p95 |
|---|---|---|---|---|---|---|---|---|---|---|---|
| ariane133@4 | 0.036 | 1.67 | 3.39 | 45.97 | 94.1 | 35.5 | 34.2 | 41.9 | 11.42 | 11.24 | 11.36 |
| ariane133@6 | 0.092 | 0.91 | 4.23 | 9.86 | 46.0 | 86.5 | 88.0 | 110.0 | 33.58 | 32.72 | 34.45 |
| bp@4 | 0.048 | 1.48 | 4.82 | 30.50 | 100.4 | 36.9 | 40.2 | 41.4 | 8.84 | 8.87 | 9.72 |
| bp@6 | 0.700 | 1.43 | 15.85 | 2.04 | 22.6 | 162.9 | 174.9 | 233.8 | 50.33 | 51.70 | 60.67 |
| bp_be@4 | 0.042 | 0.54 | 2.43 | 12.86 | 57.8 | 36.5 | 47.1 | 43.8 | 14.84 | 14.28 | 15.67 |
| bp_be@6 | 0.449 | 0.46 | 3.71 | 1.03 | 8.3 | 141.4 | 138.1 | 180.2 | 87.43 | 82.75 | 88.68 |
| swerv_wrapper@4 | 0.063 | 0.78 | 3.55 | 12.35 | 56.3 | 50.1 | 48.9 | 59.9 | 13.16 | 12.96 | 15.09 |
| swerv_wrapper@6 | 2.050 | 1.80 | 18.58 | 0.88 | 9.1 | 205.7 | 224.6 | 262.1 | 103.91 | 97.49 | 131.41 |
| ariane136@4 | 0.028 | 1.60 | 3.09 | 57.71 | 110.4 | 23.8 | 23.8 | 23.8 | 7.85 | 7.85 | 7.95 |
| ariane136@6 | 0.084 | 1.18 | 4.82 | 14.13 | 57.4 | 72.3 | 74.2 | 83.7 | 29.69 | 29.15 | 29.95 |
| bp_fe@4 | 0.199 | 0.41 | 3.26 | 2.07 | 16.4 | 71.6 | 72.6 | 80.7 | 42.89 | 42.35 | 45.81 |
| bp_fe@6 | 1.156 | 0.76 | 5.51 | 0.66 | 4.8 | 188.4 | 187.6 | 209.4 | 130.77 | 125.33 | 135.89 |
| bp_multi@4 | 0.081 | 0.90 | 5.12 | 11.03 | 63.2 | 37.9 | 35.4 | 44.1 | 14.58 | 14.53 | 16.07 |
| bp_multi@6 | 1.098 | 0.99 | 15.90 | 0.90 | 14.5 | 129.5 | 131.1 | 199.8 | 70.77 | 69.88 | 91.42 |
| bp_quad@4 | 0.654 | 12.03 | 37.44 | 18.38 | 57.3 | 70.9 | 73.0 | 77.4 | 20.26 | 20.10 | 20.64 |
| bp_quad@6 | 2.577 | 6.36 | 67.38 | 2.47 | 26.1 | 140.6 | 155.6 | 144.6 | 61.61 | 61.66 | 65.97 |
| aes@2 | 0.005 | 0.23 | 0.39 | 45.82 | 78.0 | – | – | – | – | – | – |
| dynamic_node@2 | 0.015 | 0.17 | 0.60 | 11.54 | 39.7 | – | – | – | – | – | – |
| ibex@2 | 0.005 | 0.15 | 0.29 | 32.18 | 59.0 | – | – | – | – | – | – |
| jpeg@2 | 0.042 | 0.64 | 6.01 | 15.18 | 143.1 | – | – | – | – | – | – |
| swerv@2 | 0.038 | 0.84 | 5.04 | 22.23 | 132.8 | – | – | – | – | – | – |

**Table B3.** bp pitch sweep: utilization (%), GEO-Exact optimum (not dual-certified; checked against an independent sparse exact reference) and SW total (µm), SW gap (%).

| Pitch (µm) | Util. | Optimum | SW total | SW gap |
|---|---|---|---|---|
| 6.9 | 46.1 | 217,352.89 | 217,933.95 | 0.27 |
| 7.3 | 51.5 | 294,319.77 | 295,454.35 | 0.39 |
| 7.8 | 58.5 | 405,974.33 | 431,426.35 | 6.27 |
| 8.2 | 64.5 | 504,198.87 | 610,756.17 | 21.13 |
| 8.5 | 69.5 | 582,510.19 | 640,505.65 | 9.96 |
| 8.9 | 76.7 | 699,871.47 | 723,306.15 | 3.35 |
| 9.2 | 81.5 | 800,493.65 | 825,288.57 | 3.10 |
| 9.6 | 88.6 | 956,436.41 | 957,884.95 | 0.15 |
| 9.8 | 92.6 | 1,046,593.99 | 1,046,593.99 | 0.00 |
| 9.9 | 94.7 | 1,097,390.39 | 1,097,390.39 | 0.00 |